\documentclass[10pt,twocolumn,letterpaper]{article}

\usepackage{cvpr}
\usepackage{times}
\usepackage{epsfig}
\usepackage{graphicx}
\usepackage{amsmath}
\usepackage{amssymb}
\usepackage{multirow}
\usepackage{bbm}
\usepackage{color}

\cvprfinalcopy

\ifcvprfinal\fi
\begin{document}

\title{Learning Multi-level Region Consistency with Dense Multi-label Networks \\
for Semantic Segmentation}

\author{Tong Shen{$^{*}$}, ~ Guosheng Lin\thanks{The first two authors contributed equally. Correspondence should be addressed to C. Shen.}, ~ Chunhua Shen, ~ Ian Reid\\
	School of Computer Science, The University of Adelaide, Adelaide, SA 5005, Australia\\
	e-mail: \small \tt firstname.lastname@adelaide.edu.au}

\maketitle

\begin{abstract}
	Semantic image segmentation is a fundamental task in image understanding. Per-pixel semantic labelling of an image benefits greatly from the ability to consider region consistency both locally and globally. However, many Fully Convolutional Network based methods do not impose such consistency, which may give rise to noisy and implausible predictions. We address this issue by proposing a dense multi-label network module that is able to encourage the region consistency at different levels. This simple but effective module can be easily integrated into any semantic segmentation systems. With comprehensive experiments, we show that the dense multi-label can successfully remove the implausible labels and clear the confusion so as to boost the performance of semantic segmentation systems.
\end{abstract}

\tableofcontents
\clearpage

\section{Introduction}
\begin{figure}[t!]
\begin{center}
   \includegraphics[width=1.0\linewidth]{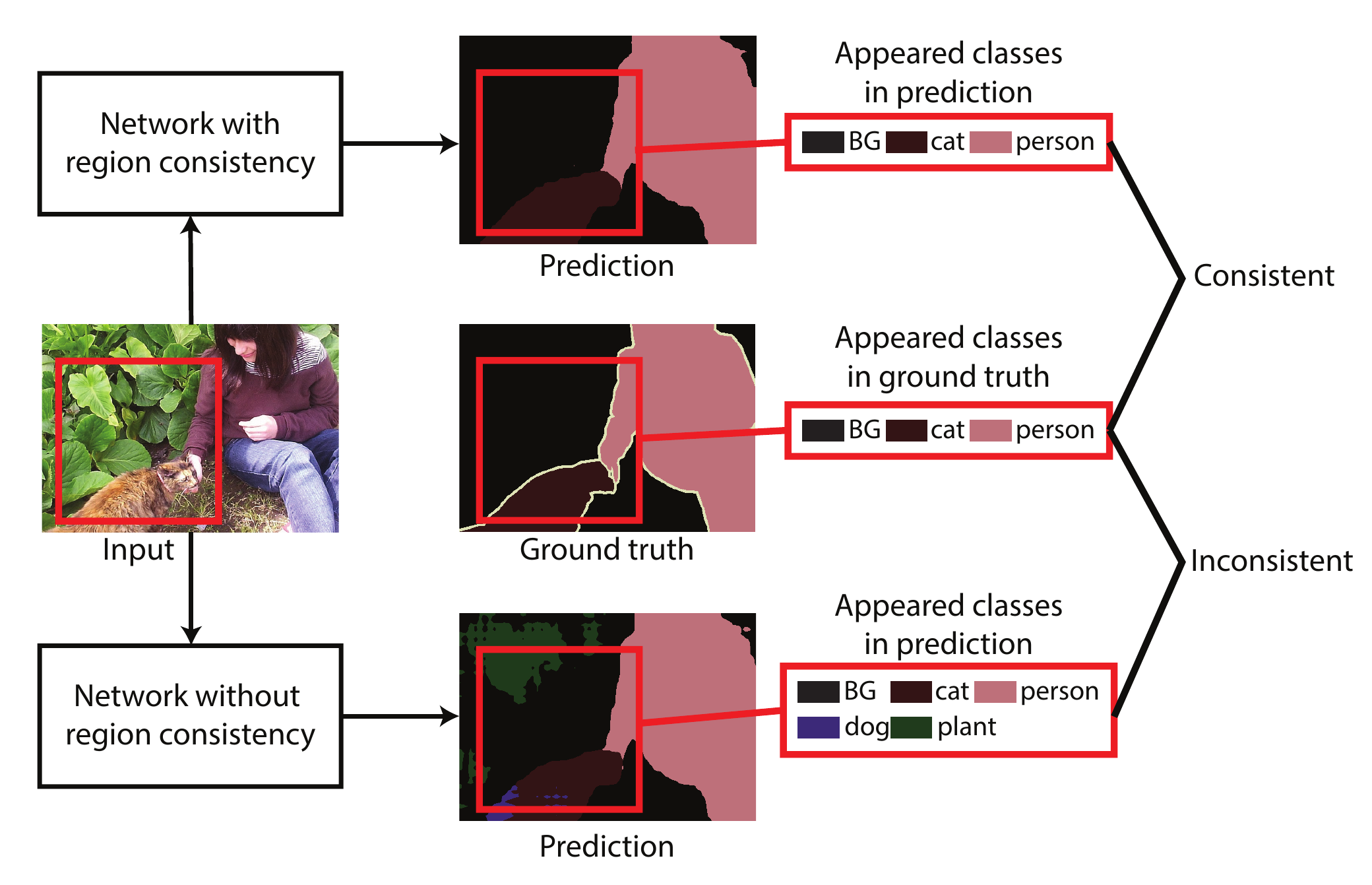}
\end{center}
   \caption{Illustration of region consistency. For a region in the input image, which is coloured in red, the corresponding part in the ground truth contains only three classes. In the network without region consistency, there are five classes that appear. If we explicitly encourage the consistency, those unlikely classes will be eliminated and the prediction will be better as shown on top.}
\label{fig:consistency}
\end{figure}

Semantic segmentation is one of the fundamental problems in computer vision, whose task is to assign a semantic label to each pixel of an image so that different classes can be distinguished. This topic has been widely studied \cite{Girshick2014, Carreira2012, Hariharan2014, Yadollahpour2013, Farabet2013, Cogswell2014}. Among these models, Fully Convolutional Network (FCN) based models have become dominant \cite{Dai2015, Hong, Shen2016, Chen2014a, Lin2015, Chen2016}. These models are simple and effective because of the powerful capacity of Convolutional Neural Networks (CNNs) and being able to be trained end-to-end. However, most existing methods do not have the mechanism to enforce the region consistency, which plays an important role in semantic segmentation. Consider, for example, Figure \ref{fig:consistency}, in which the lower left image is the output of a vanilla FCN, whose prediction contains some noisy labels that do not appear in the ground truth. With enforced region consistency, we can simply eliminate those implausible labels and clear the confusion. Our aim in this work is to introduce constraints to encourage this consistency. 

Our proposal is both simple and effective: we argue that the region consistency in a certain region can be formulated as a multi-label classification problem.  Multi-label classification has also been widely studied \cite{Wei2014, Jiang2016, Moore2016, Guo2011, Gong2013}, whose task is to assign one or more labels to the image. By performing multi-label classification in a region, we can allow the data to suggest which labels are likely within the broad context of the region, and use this information to suppress implausable classes predicted without reference to the broader context, thereby improving scene consistency. While typical multi-label problems are formulated as whole-image inference, we adapt this approach to dense prediction problems such as semantic segmentation, by introducing dense multi-label prediction for image regions of various sizes. 

\begin{figure}[t!]
\begin{center}
   \includegraphics[width=1.0\linewidth]{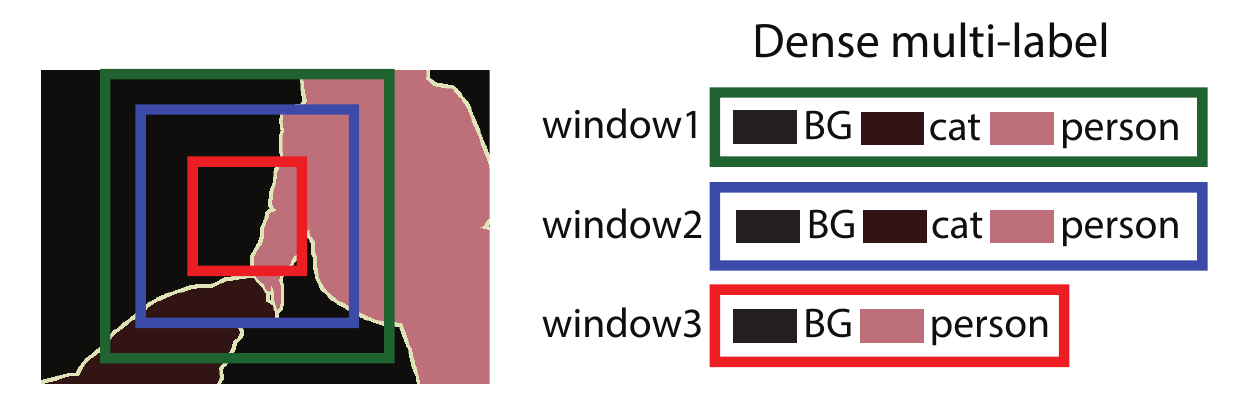}
\end{center}
   \caption{Illustration of dense multi-label with multi-level. There are three windows with different sizes. The red window, the smallest, focuses more on the local region consistency, while the green window is responsible for global region consistency. The other one, in blue, is for mid-level consistency. By sliding the windows, we can perform multi-label densely for each spatial point.}
\label{fig:dense_multi_illu}
\end{figure}

Dense multi-label prediction is performed in a sliding window fashion: the classification for each spatial point is influenced by the network prediction and by the multi-label result for the surrounding window. By employing different window sizes, we are able to construct a multi-level structure for dense multi-label and enforce the region consistency at different levels both locally and globally. Figure \ref{fig:dense_multi_illu} is an illustration of dense multi-label at multiple windows sizes.Here we use three windows of different sizes. The red window, the smallest, focuses more on the local region consistency, while the green window, the largest, is responsible for global region consistency. The other one, in blue, is for mid-level consistency. By sliding the windows to consider each spatial point, we perform multi-label densely at different level, encouraging the segmentation predictor to give predictions that are consistent with the dense multi-label prediction.

Our contributions are as follows:

\begin{itemize}
\item We address the problem of region consistency in semantic segmentation by proposing a dense multi-label module to achieve the goal of retaining region consistency, which is simple and effective. We also introduce a multi-level structure for dense multi-label to preserve region consistency both locally and globally.

\item We evaluate our method on four popular semantic segmentation datasets including NYUDv2, SUN-RGBD, PASCAL-Context and ADE 20k, and achieve promising results. We also give analysis on how dense multi-label can remove the implausible labels, clear confusion and effectively boost the segmentation systems.  
\end{itemize}

This paper is organized as follows. Firstly we review related work in Section \ref{related_work}. We then explain dense multi-label and describe the overview of our structure in Section \ref{methods}. In Section \ref{exp}, we show comprehensive experiments and analyze the results. In the end, we draw conlusions and talk about future work in Section \ref{conclusion}.

\section{Related Work}
\label{related_work}

Semantic segmentation has been widely studied \cite{Girshick2014, Carreira2012, Hariharan2014, Yadollahpour2013, Farabet2013, Cogswell2014}. Early CNN based methods rely on region proposals or superpixels. They make segmentation prediction by classifying these local features.

More recently, with Long \etal\cite{Long2015} introducing applying Fully Convolutional Networks(FCNs) to semantic segmentation, the FCN based segmentation models \cite{Dai2015, Hong, Shen2016, Chen2014a, Lin2015, Chen2016} have become popular. In \cite{Long2015}, Long \etal convert the last fully connected layers into convolutional layers thus make the CNN accept abitrary input size. Since the output retains the spatial information, it is straightforward to train the network jointly in an end-to-end fashion. They also introduce skip architecture to combine features from different levels. Chen \etal \cite{Chen2014a} modify the original FCN by introducing dilated kernels, in which kernels are inserted with zeros, to enable large field of view and Fully Connected CRF to refine outputs. Lin \etal \cite{Lin2015} introduces a joint training model with CRFs. In this work, CRFs are not simply used for smoothness as in \cite{Chen2014a}, but  a more general term to learn context information to help boost the unary performance. Liu \etal \cite{Liu2015b} utilise global features to improve semantic segmentation. They extract global features from different levels and fuse them by using L2 normalization layer. Our method is different from those. We attempt to improve the performance of segmentation by enforcing region consistency using dense multi-label.

Multi-label classification has also been widely studied. Traditional methods are based on graphical models \cite{Xue2011, Guo2011}, while the recent studies benefit more from CNNs \cite{Wei2014, Jiang2016, Gong2013}. Gong \etal \cite{Gong2013} transform a single-label classification model into multi-label classification model and use ranking loss to train the model. Wei \etal \cite{Wei2014} also use the transfer learning from single-label classification models. They perform the multi-label classification by first generating the object hypotheses and the fusing predictions as the final prediction for the whole image. Jiang \etal \cite{Jiang2016} propose a unified framework for multi-label classification by using CNN and Recurrent Neural Network (RNN).

Here we propose a dense multi-label module to take advantage of multi-label classification and integrate it into semantic segmentation systems. Dense multi-label is performed in a sliding window fashion and treats all area in a window as multi-label classification. Experiments show that dense multi-label can help to keep the scene consistency, clear confusion and boost the performance of semantic segmentation.

\section{Methods}
\label{methods}
\subsection{Dense Multi-label}
\begin{figure*}[h]
\begin{center}
  \includegraphics[width=.8\linewidth]{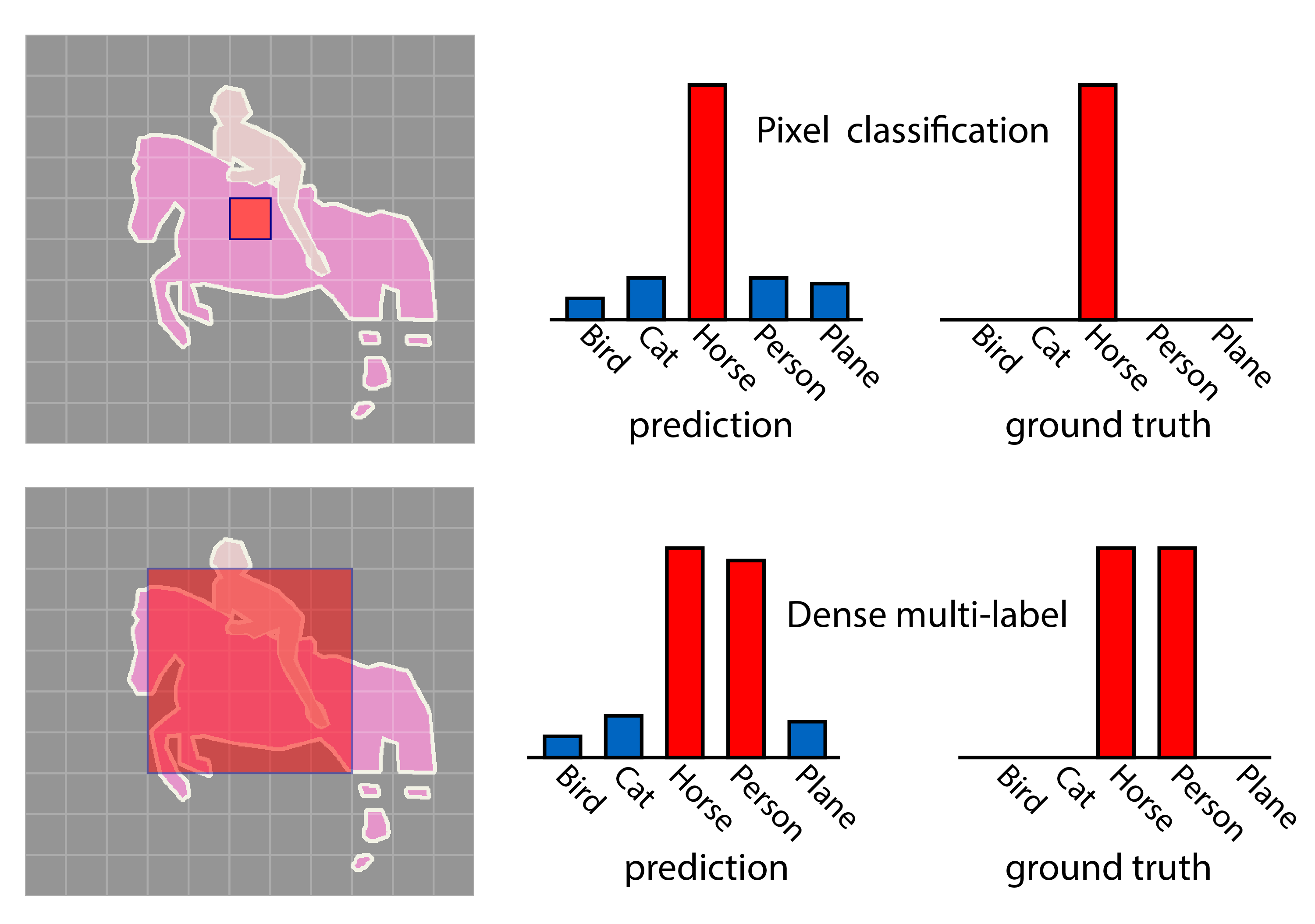}
\end{center}
   \caption{An illustration of differences between pixel classification and dense multi-label prediction. In pixel classification, we treat each spatial point as a single-label classification problem where only one class is supposed to get very high confidence; dense multi-label focuses on label concurrence where the labels that appear in the region will have equally high confidence. }
\label{fig:multi_seg}
\end{figure*}

Multi-label classification is a task where each image can have more than one label, unlike a multi-class classification problem \cite{Arge2015, Szegedy, Technologii2013, Krizhevsky2012} whose goal is to assign only one label to the image. This is more natural in reality because for majority of images, objects are not isolated, instead they are in context with other objects or the scene. Multi-label classification gives us more information of the image.

For a dense prediction task such as segmentation, it treats every spatial point as a multi-class classification problem, where the point is assigned with one of the categories. As shown in the upper part of Figure \ref{fig:multi_seg}, the model predicts scores for each class and picks the highest one. The ground truth is an one-hot vector correspondingly. For a dense multi-label problem, each spatial point will be assigned with several labels to show what labels appear in the a certain window centered at this point. As shown in lower part of Figure \ref{fig:multi_seg}, there are two classes being predicted with high confidence and the ground truth is given by a ``multiple hot'' vector.

Here we propose a method to learn a dense multi-label system and a segmentation system at the same time. We aim at using dense multi-label to suppress the implausible classes and encourage appropriate classes so as to retain the region consistency for the segmentation prediction both globally and locally. In the next section, more details of the whole framework will be provided.

\subsection{Overview of Framework}
\begin{figure*}[t!]
\begin{center}
   \includegraphics[width=0.9\linewidth]{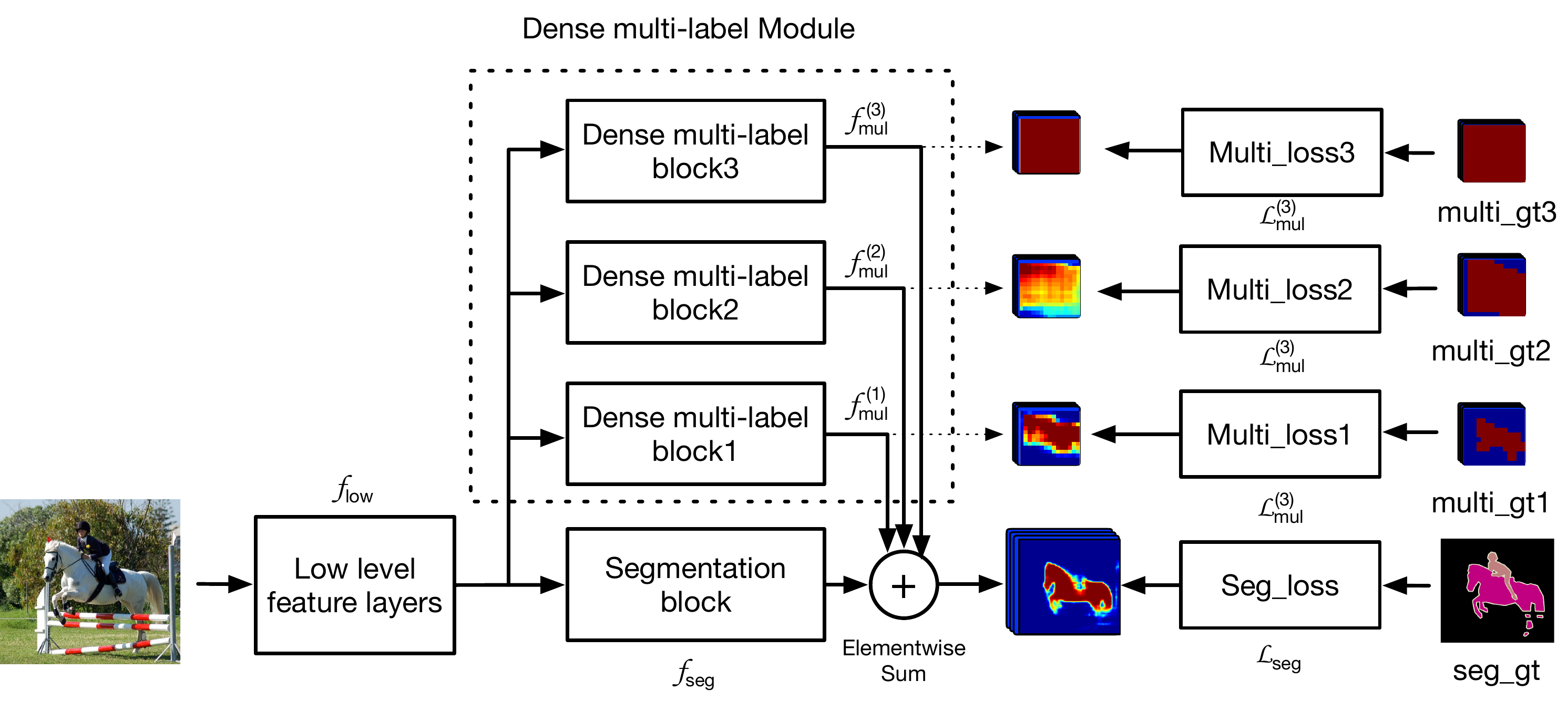}
\end{center}
   \caption{Illustration of the framework with dense multi-label module. The input image is first passed into low level feature layers, which are shared by the following blocks. Then the feature maps are fed into the segmentation block and three dense multi-label blocks. The element-wise sum will sum up the features from the blocks and make the final prediction. Apart from the segmentation loss, each dense multi-label block also has its own multi-label loss to guide the training.}
\label{fig:pipeline}
\end{figure*}

An  overview of the structure is shown in Figure \ref{fig:pipeline}, with the part in the dashed-line rectangle being the dense multi-label module. Without it, the network simply becomes a FCN. The input image is first fed into several low level feature layers which are shared by the following blocks. Then apart from going into the segmentation block, the features also enter three blocks for dense multi-label prediction. The outputs of theses blocks are merged element-wise to make the final prediction.

In the training phase, the network is guided by four loss functions: the segmentation loss and three dense multi-label losses. We use softmax loss for the segmentation path, and use logistic loss for all the dense multi-label blocks.

The dense multi-label blocks have different window sizes for performing dense multi-label prediction within different contexts. With this multi-level structure, we are able to retain region consistency both locally and globally.

Let $\boldsymbol{x}$ denote the image. The process of the low level feature block can be described as:
\begin{equation}
 \boldsymbol{o} = f_{low}( \boldsymbol{x}; \boldsymbol{\theta}_{low}),
\end{equation}
where $\boldsymbol{o}$ is the output and $\boldsymbol{\theta}_{low}$ the layer parameters.

The dense multi-label blocks and the segmentation block are defined as:
\begin{equation}
 \boldsymbol{m}^{(j)} = f_{mul}^{(j)}(\boldsymbol{o}; \boldsymbol{\theta}_{mul}^{(j)}), j\in\{1,2,3\}
\end{equation}
\begin{equation}
\boldsymbol{s}=f_{seg}(\boldsymbol{o};\boldsymbol{\theta}_{seg}),
\end{equation}
where $\boldsymbol{m}^{(j)}$ and $\boldsymbol{s}$ denote the output of $j$th multi-label block and the output of segmentation respectively. $\boldsymbol{\theta}_{mul}^{(j)}$ and  $\boldsymbol{\theta}_{seg}$ are layer parameters.

The final prediction is:
\begin{equation}
\boldsymbol{p}=\boldsymbol{s}+\boldsymbol{m}^{(1)}+\boldsymbol{m}^{(2)}+\boldsymbol{m}^{(3)},
\end{equation}
where $\boldsymbol{p}$ is the fused score for segmentation.

For the loss functions, we use logistic loss for the prediction of dense multi label blocks, $\boldsymbol{m}^{(1)}$,$\boldsymbol{m}^{(2)}$ and $\boldsymbol{m}^{(3)}$; softmax loss is used for final prediction $\boldsymbol{p}$. Let $m_{ik}$ be the out of a dense
multi-label block at $i$th position for $k$th class, and $y_{ik}^{mul}$
be the ground truth for the corresponding position and class.
The loss function for dense multi-label is defined as:

\begin{align}
l_{mul}(\boldsymbol{y}^{mul},\boldsymbol{m}) & =\frac{1}{IK}\sum_{i}^{I}\sum_{k}^{K}y_{ik}^{mul}\log(\frac{1}{1+e^{-m_{ik}}})\nonumber\\
& +(1-y_{ik}^{mul})\log(\frac{e^{-m_{ik}}}{1+e^{-m_{ik}}}),
\end{align}
where $y_{ik}^{mul}\in\{0,1\}$; $I$ and $K$ represent the number of spatial points and classes, respectively.

Similarly, let $p_{ik}$ be the fused output at $i$th position for $k$th class, and $y_{i}^{seg}$ be the ground truth
for segmentation prediction at $i$th position. The loss function for segmentation is defined as:

\begin{equation}
l_{seg}(\boldsymbol{y}^{seg},\boldsymbol{p})=\frac{1}{I}\sum_{i}^{I}\sum_{k}^{K}\mathbbm{1}(y_{i}^{seg}=k)\log(\frac{e^{p_{ik}}}{\sum_{j}e^{p_{ij}}}),
\end{equation}

where $y_{i}^{seg}\in\{1 \dots K\}$.

Our goal is to minimize the objective function:
\begin{equation}
\min l_{seg}+\lambda(l_{mul}^{(1)}+l_{mul}^{(2)}+l_{mul}^{(3)}),
\end{equation}
where $\lambda$ controls the balance between the segmentation block and the dense multi-label blocks. I observe this parameter is not very sensitive. We set $\lambda=1$ to treat each part equally.

\subsection{Dense Multi-label Block}
\label{sec:densemultilabel}
\begin{figure}[h]
\begin{center}
   \includegraphics[width=0.9\linewidth]{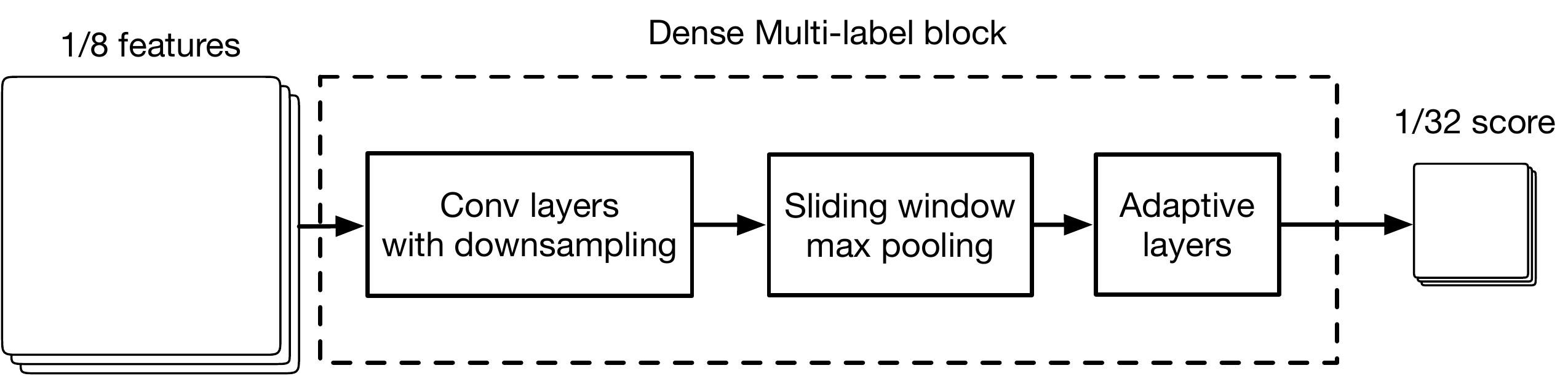}
\end{center}
   \caption{Details of a single dense multi-label block. The input features are fed into several convolutional layers and further downsampled. Then we perform sliding window with max pooling operation. After some adaptive layers, we have scores for dense multi-label at 1/32 resolution.}
\label{fig:dense_multi_block}
\end{figure}

The details of the dense multi-label block are shown in Figure \ref{fig:dense_multi_block}, where the input is feature maps at 1/8 resolution, due to the downsampling in the low level feature layers. After some convolutional layers with further downsampling, the dense multi-label is performed at 1/32 resolution with the sliding window and following adaptive layers. The reason for this setting is because dense multi-label requires a large sliding window, which will become a computational burden if we work at a high resolution. Downsampling can greatly reduce the size of feature maps and more importantly, the size of sliding window will shrink accordingly, thus making the computation more efficient. On the other hand, dense multi-label requires more high level information. Therefore, working at a coarse level can capture the high level features better. The output of the dense multi-label is upsampled to be compatible with the segmentation block's output.

\subsection{Ground Truth Generation}

\begin{figure}[h]
\begin{center}
   \includegraphics[width=0.9\linewidth]{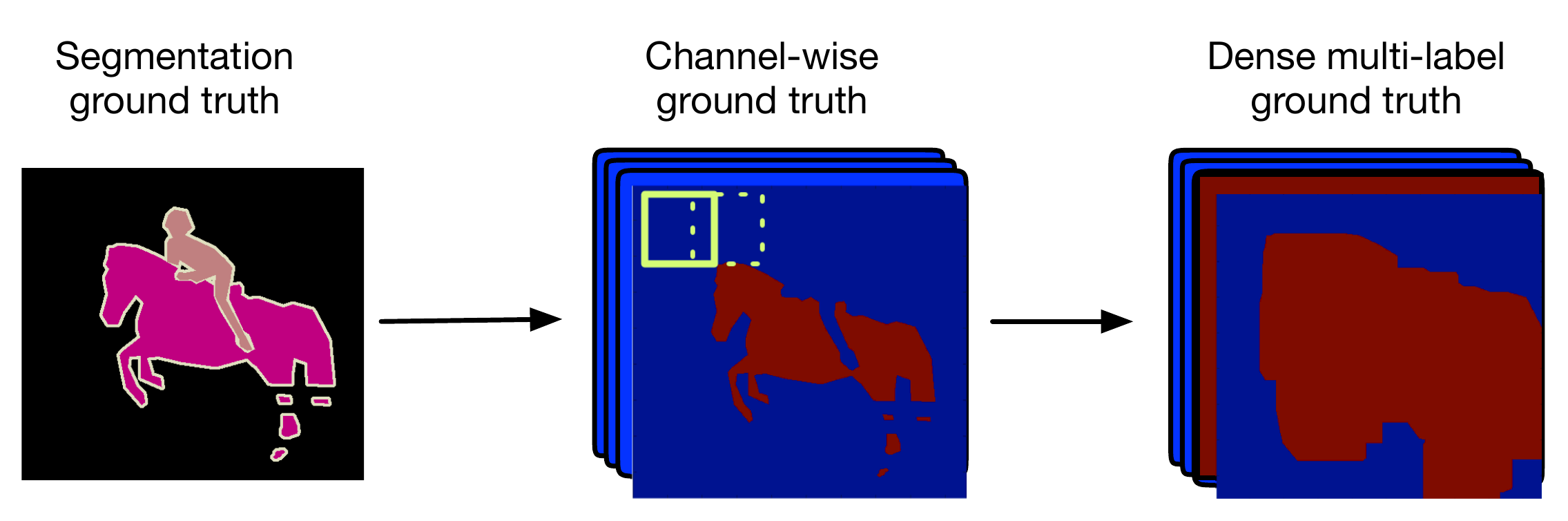}
\end{center}
   \caption{The segmentation ground truth is firstly converted to channel-wise labels, with 0 or 1 in each channel. The ground truth for dense multi-label can be obtained by performing max pooling on the channel-wise labels.}
\label{fig:gt_generation}
\end{figure}

The ground truth for dense multi-label can be generated from the segmentation ground truth. The process is described in Figure \ref{fig:gt_generation}. Firstly, the segmentation ground truth is converted to channel-wise labels, which means each channel only contains 1 or 0 to indicate whether the corresponding class appears or not. To generate a ground-truth mask for each class, for a given window size, we slide the window across each binary channel and perform a max-pool operation (this is equivalent to a binary dilation using a structuring element of the same size and shape as the window).  We repeat this process for each window size. As noted in section \ref{sec:densemultilabel}, the dense multi-label classification is performed at 1/32 resolution while the segmentation is at 1/8. Therefore, we generate multi-label ground-truth data at 1/8 resolution with stride 4.

\subsection{Network Configuration} \label{netconf}
The dense multi-label module is suitable for any segmentation system and it can be easily integrated. In this study, we use Residual 50-layer network \cite{Technologii2013} with dilated kernels \cite{Chen2014a}.

In order to work at a relatively high resolution while keeping the efficiency, we use 8-stride setting, which means that the final output is at 1/8 resolution. As we mentioned in the last section, we perform dense multi-label at 1/32 resolution to make it more efficient and effective. The window sizes are then defined at 1/32 resolution. For example,let $w$ be the window size. A window with $w=17$ at 1/32 resolution means $4w=68$ at 1/8 resolution. The corresponding window for the original image is $32w=544$. We use $w_1=35$, $w_2=17$ and $w_3=7$ for all the experiments.

\begin{table}
\begin{centering}
\begin{tabular}{|c|c|c|}
\hline
Block name & Initial layers & Stride\tabularnewline
\hline
\hline
Low level feature block & conv1 to res3d & 8\tabularnewline
\hline
Segmentation block & res4a to res5c & 1\tabularnewline
\hline
Dense multi-label block & res4a to res5c & 4\tabularnewline
\hline
\end{tabular}
\par\end{centering}
\caption{Configuration for Res50 network. The low level feature block is initialized by layers ``conv1'' to ``res3d'' and has 8 stride. The segmentation block and dense multi-label blocks are initialized by layers ``res4a'' to ``res5c'' but do not share the weights with each other. The segmentation block does not have any downsampling, but the dense multi-label blocks have further 4 stride downsampling. }
\label{tab:res50_conf}
\end{table}

Table \ref{tab:res50_conf} shows the layer configuration with Residual network with 50 layers (Res50) as the base network. The low level feature block contains the layers from ``conv1'' to ``res3d''. The segmentation block and dense multi-label blocks have the layers from ``res4a'' to ``res5c'' as well as some adaptive layers. It is worth noting that it does not mean the segmentation block and dense multi-label blocks will share the weights even though they initialize the weights from the same layers. After initialization, they will learn their own features separately.

\section{Experiments and Analysis}
\label{exp}
We evaluate our model on 4 commonly used semantic segmentation datasets: ADE 20k, NYUDv2, SUN-RGBD and PASCAL-Context. Our comprehensive experiments show that dense multi-label can successfully suppress many unlikely labels, retain region consistency and thus improve the performance of semantic segmentation. 

The results are evaluated using the Intersection-over-Union (IoU) score \cite{Everingham2010}. Moreover, since our original motivation is to suppress noisy and unreasonable labels to keep labels consistent with the region, we also introduce new measurements to evaluate the number of classes that are not in ground truth, and further, the number of pixels that are predicted to be these wrong classes for each image. 

We only use Res50 as base network to compare and analyse the performance. For all the experiments, we use batch size of 8, momentum of 0.9 and weight decay of 0.0005. 

\subsection{Results on ADE 20k dataset}
We first evaluate our result on ADE 20k dataset\cite{Zhou2016}, which contains 150 semantic categories including objects such as person, car \etc, and ``stuff'' such as sky, road \etc. There are 20210 images in the training set and 2000 images in the validation set. 

\begin{figure}[h]
\begin{center}
   \includegraphics[width=0.9\linewidth]{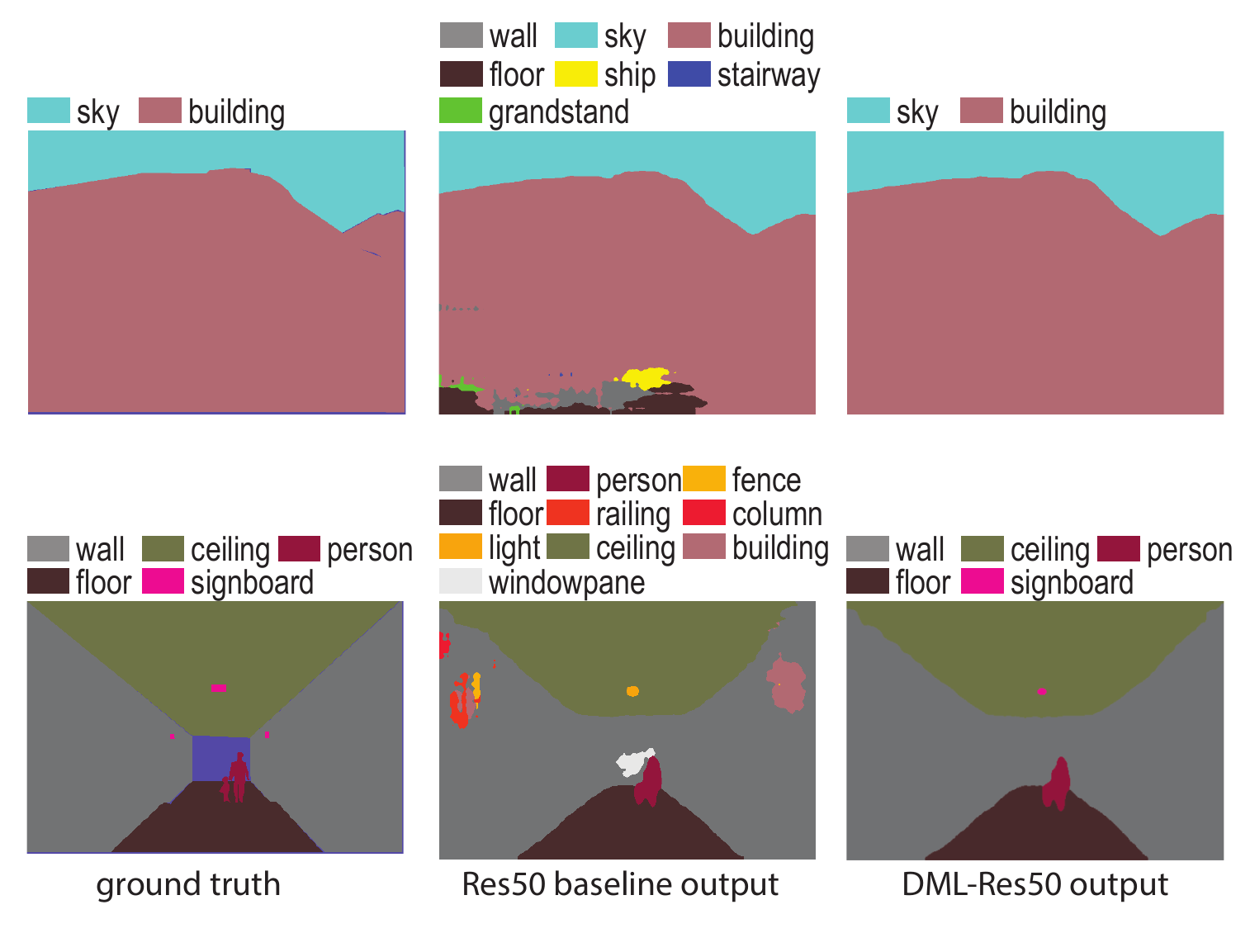}
\end{center}
   \caption{Example outputs of Res50 baseline and DML-Res50 on ADE 20k dataset.}
\label{fig:ade_exp}
\end{figure}

As shown in Table \ref{tab:ade_exp}, the model with dense multi-label (DML-Res50) yields a 2\% improvement. 
To analyse the effectiveness of label suppression, we also use two criteria to evaluate this performance, which are shown as ``Wrong class'' and ``Wrong labels''. Wrong class means the number classes that are not supposed to appear but are mistakenly predicted by the model. Wrong labels describe how many pixels are assigned with those wrong classes. We observe that using Dense multi-label effectively reduces the wrong classes and labels, by 35\% and 16\% respectively.  Some examples are shown in Figure \ref{fig:ade_exp}. To make fair comparison, all the images are raw outputs directly from the network. The last column shows the outputs from the network with dense multi-label where we can observe great scene consistency compared with the output of the baseline network shown in the middle.  

\begin{figure}[h]
\begin{center}
   \includegraphics[width=0.9\linewidth]{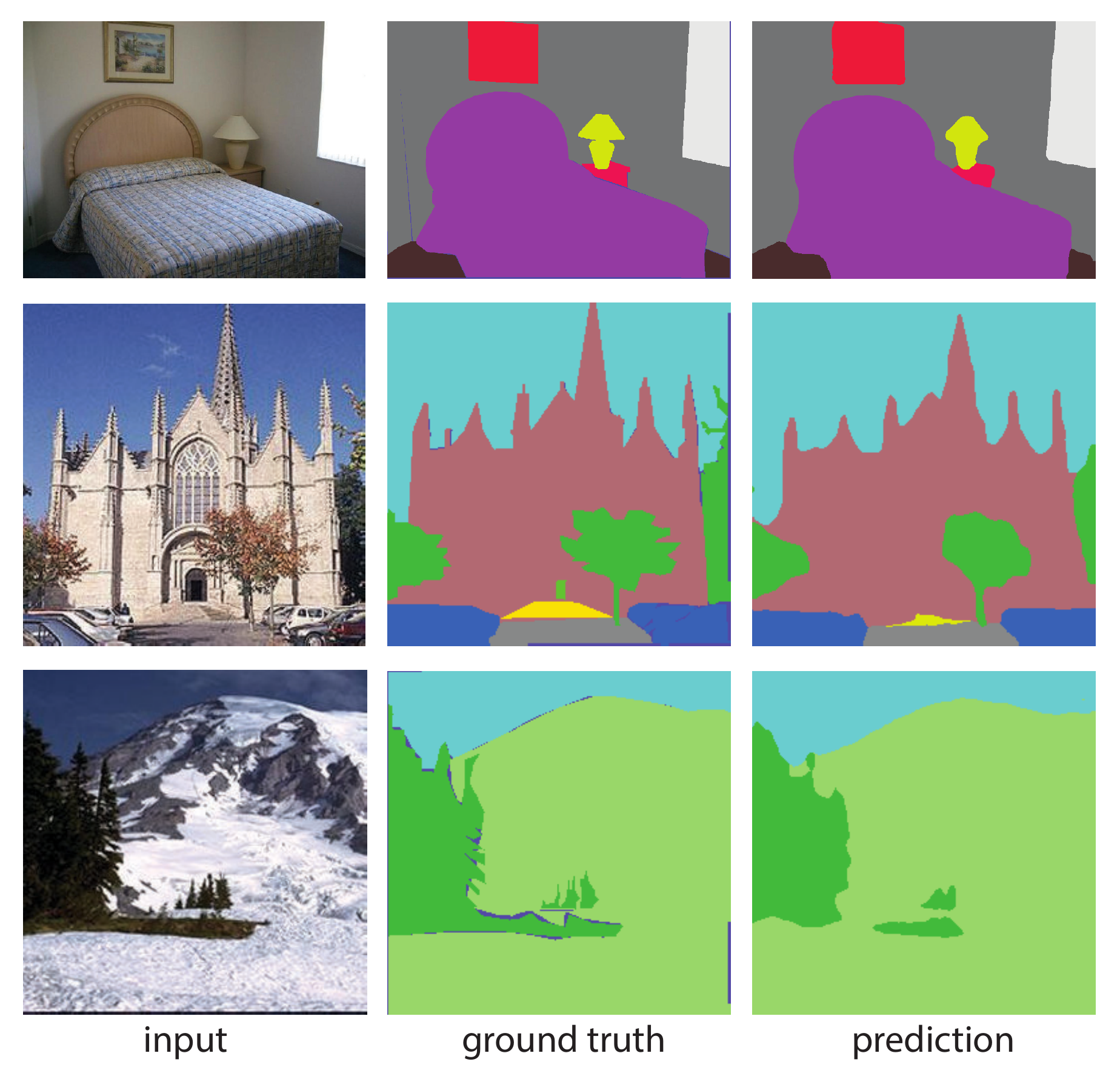}
\end{center}
   \caption{More example outputs of dense multi-label network on ADE dataset.}
\label{fig:ade_more_exp}
\end{figure}

In comparison with other methods, we achieve better results than the models reported in \cite{Zhou2016}, as shown in Table \ref{tab:ade_comp}. More examples can be found in Figure \ref{fig:ade_more_exp}

\begin{table}
\begin{centering}
\begin{tabular}{|c|c|c|c|}
\hline 
Model & IOU & \#Wrong class & \#Wrong label\tabularnewline
\hline 
\hline 
Res50 baseline & 34.5 & 5.576 & 21836\tabularnewline
\hline 
DML Res50 & \textbf{36.49} & \textbf{3.6} & \textbf{18294}\tabularnewline
\hline 
\end{tabular}
\par\end{centering}
\caption{Results on ADE dataset. The dense multi-label boosts the performance by 2\% of IOU and helps reduce the number of wrong class and label by 35\% and 16\% respectively.}

\label{tab:ade_exp}
\end{table}

\begin{table}
\begin{centering}
\begin{tabular}{|c|c|}
\hline 
Model & IOU\tabularnewline
\hline 
\hline 
DilatedNet \cite{Zhou2016} & 32.31\tabularnewline
\hline 
Cascade-DilatedNet \cite{Zhou2016} & 34.90\tabularnewline
\hline 
DML-Res50(ours)  & \textbf{36.49}\tabularnewline
\hline 
\end{tabular}
\par\end{centering}
\caption{Comparsion with other models on ADE dataset. Our model achieves the best performance.}

\label{tab:ade_comp}
\end{table}

\subsection{Results on PASCAL-Context}
PASCAL-Context dataset \cite{Mottaghi2010} is a set of additional annotations for PASCAL VOC 2010, which provides annotations for the whole scene with 60 classes (59 classes and a background class). It contains 4998 images in training set and 5105 images in validation set. 

\begin{figure}[h]
\begin{center}
   \includegraphics[width=0.9\linewidth]{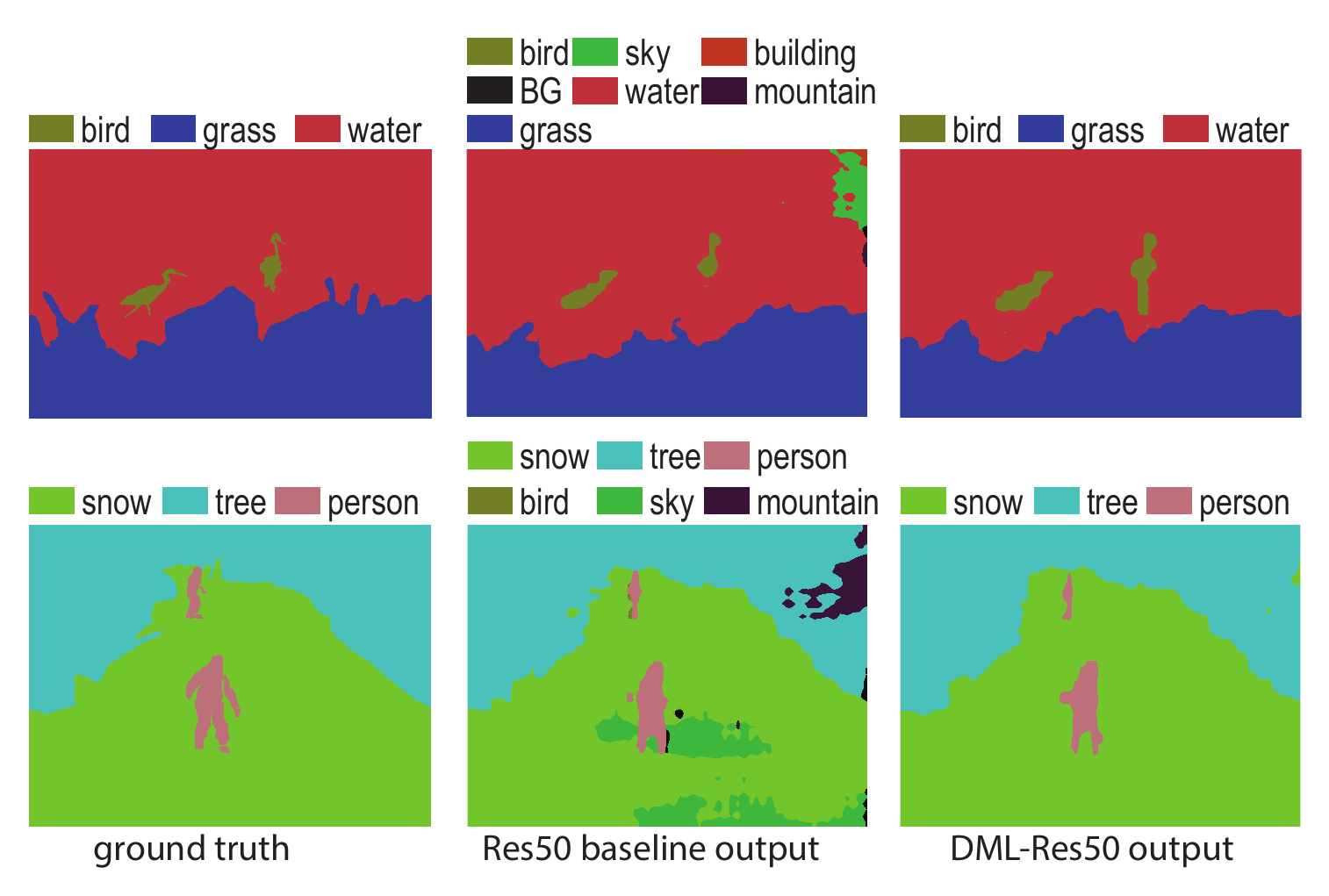}
\end{center}
   \caption{Example outputs of Res50 baseline and DML-Res50 on PASCAL-Context dataset.}
\label{fig:context_exp}
\end{figure}

Figure \ref{fig:context_exp} shows some typical examples on this dataset. We can also see clear scene consistency with dense multi-label involved. The outputs in the middle contain many noisy classes, especially the lower middle image contains ``bird'' and ``sky'', which are very unlikely in this scene. From Table \ref{tab:context_exp}, we can also see the great boost with dense multi-label. The wrong classes and labels are greatly reduced by 37\% and 15\%.

\begin{figure}[h]
\begin{center}
   \includegraphics[width=0.9\linewidth]{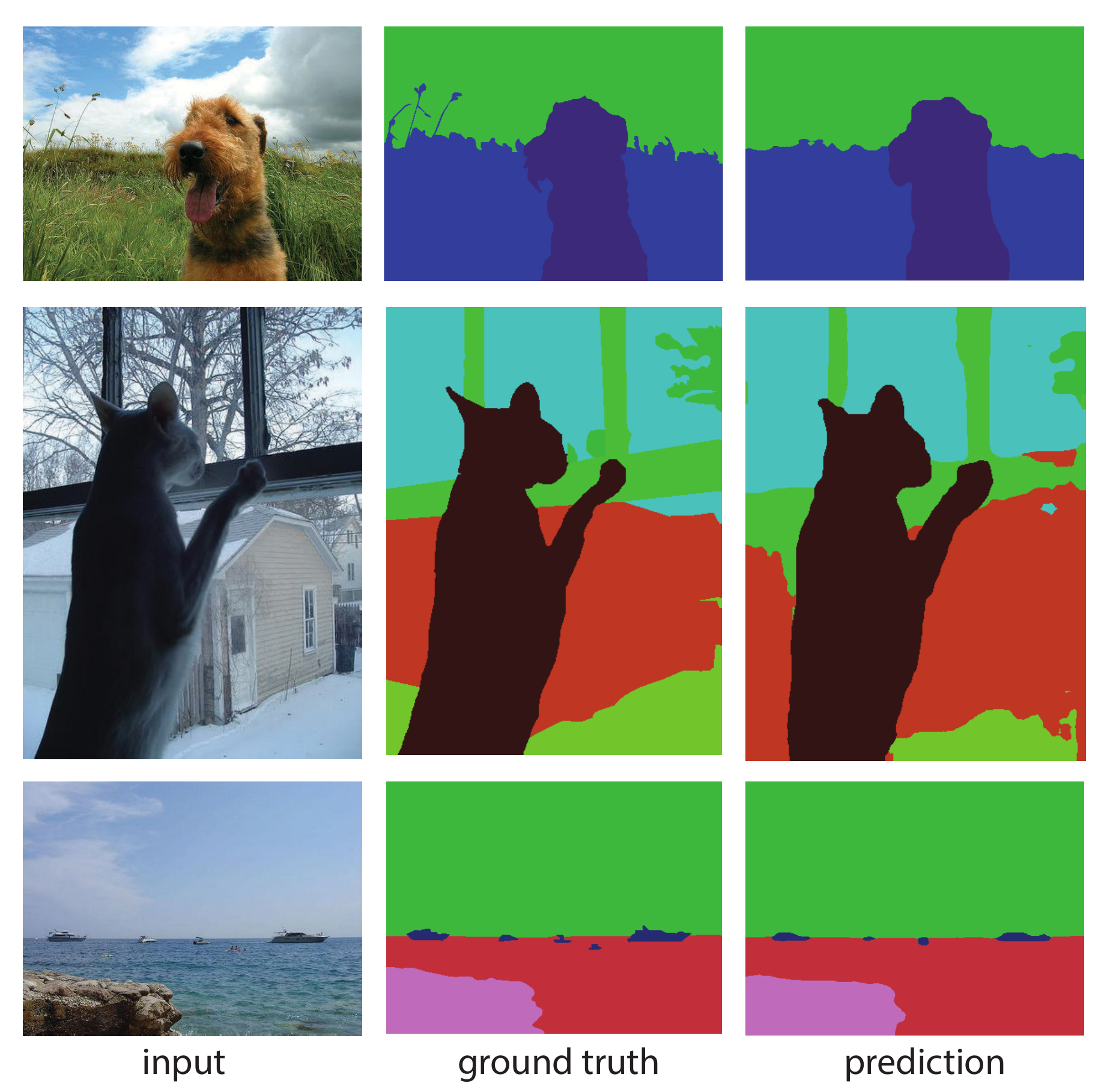}
\end{center}
   \caption{More example outputs of dense multi-label network on PASCAL-Context dataset.}
\label{fig:context_more_exp}
\end{figure}

To compare with other models, we list several results on this dataset. Since different models have various settings such as multi-scale training, extra data, \etc we also explain it in Table \ref{tab:context_comp}. Considering all the factors involved, our method is comparable since we only use Res50 as the base network and do not use mult-scale training and extra MS-COCO data for pretraining. More examples are shown in \ref{fig:context_more_exp}.
\begin{table}
\begin{centering}
\begin{tabular}{|c|c|c|c|}
\hline 
Model & IOU & \#Wrong class & \#Wrong label\tabularnewline
\hline 
\hline 
Res50 baseline & 41.37 & 4.5 & 26308\tabularnewline
\hline 
DML-Res50 & \textbf{44.39} & \textbf{2.8} & \textbf{22367}\tabularnewline
\hline 
\end{tabular}
\par\end{centering}
\caption{Results on PASCAL-Context dataset. The dense multi-label model increases the IOU by 3\% and reduces the wrong classes and labels by 37\% and 15\%. }

\label{tab:context_exp}
\end{table}

\begin{table}
\begin{centering}
\begin{tabular}{|c|c|c|c|c|}
\hline 
Model & Base& MS & Ex data &IOU\tabularnewline
\hline 
\hline 
FCN-8s \cite{Long2015} & VGG16 & no & no & 37.8\tabularnewline
\hline 
PaserNet \cite{Liu2015b} & VGG16 & no & no & 40.4\tabularnewline
\hline 
HO\_CRF \cite{Arnab2015} & VGG16 & no & no & 41.3\tabularnewline
\hline 
Context \cite{Lin2016} & VGG16 & yes & no & 43.3\tabularnewline
\hline 
VeryDeep \cite{Wu2016b} & Res101 & no & no & 44.5\tabularnewline
\hline 
DeepLab \cite{Chen2016} & Res101 & yes & COCO & \textbf{45.7}\tabularnewline
\hline
DML-Res50 (ours) & Res50 & no & no & 44.39\tabularnewline
\hline 
\end{tabular}
\par\end{centering}
\caption{Results on PASCAL-Context dataset. MS means using multi-scale inputs and fusing the results in training. Ex data stands for using extra data such as MS-COCO \cite{Lin2014}. Compared with state of the art, since we only use Res50 instead of Res101 and do not use multi-scale training as well as extra data, our result is comparable.}

\label{tab:context_comp}
\end{table}

\subsection{Results on NYUDv2}
NYUDv2 \cite{Silberman2012} is comprised of 1449 images from a variety of indoor scenes. We use the standard split of 795 training images and 654 testing images.

\begin{figure}[h]
\begin{center}
   \includegraphics[width=0.9\linewidth]{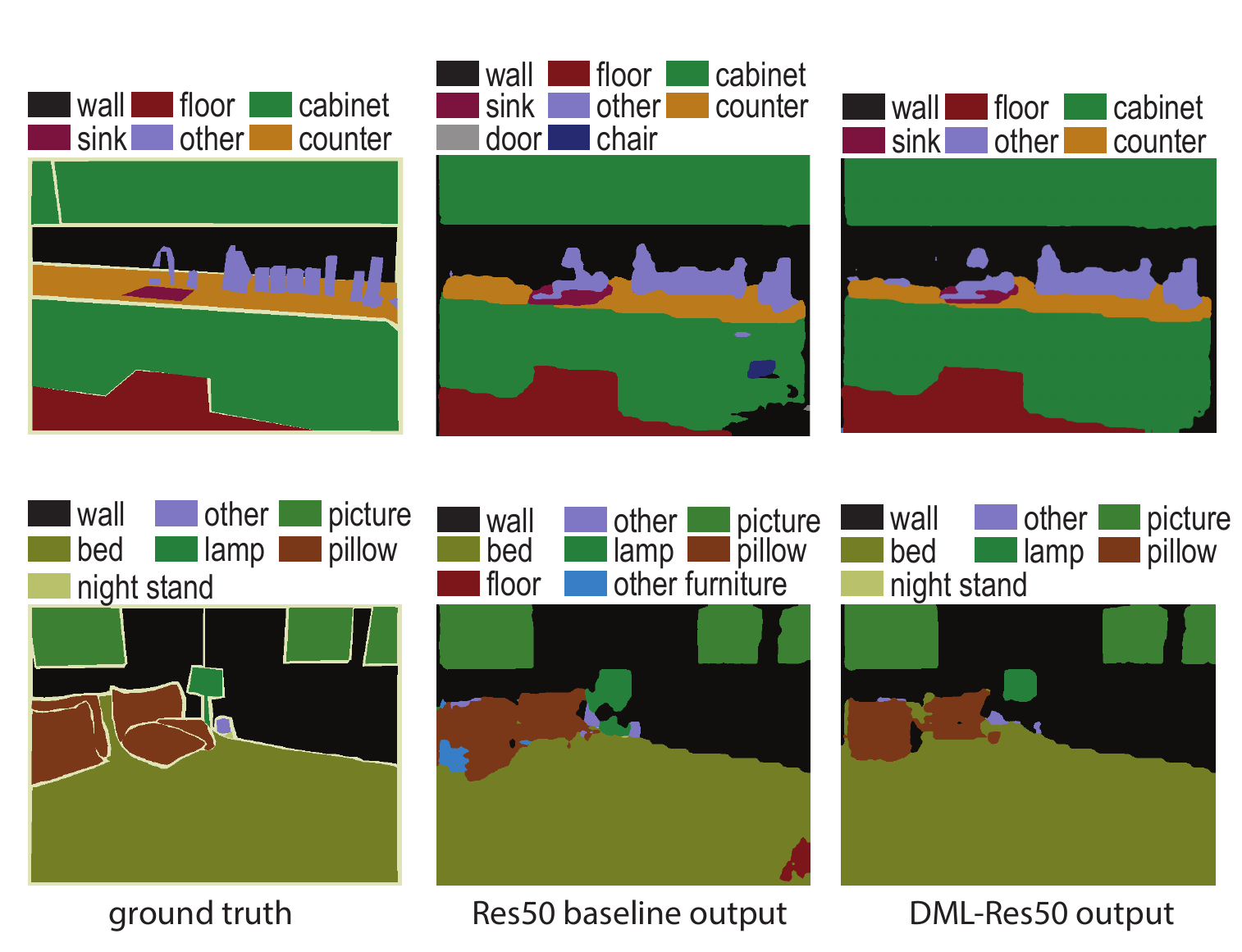}
\end{center}
   \caption{Example outputs of Res50 baseline and DML-Res50 on NYUDv2 dataset.}
\label{fig:nyud_exp}
\end{figure}

Table \ref{tab:nyud_exp} shows the results on this dataset. With dense multi-label, the performance is improved by more than 1\%, and the number of wrong class and label decrease by about 40\% and 16\%. Some examples are shown in Figure \ref{fig:nyud_exp}. Scene consistency still plays an important role in removing those noisy labels. Compared with some other models, we achieve the best result, as shown in Table \ref{tab:nyud_comp}.

\begin{table}[h]
\begin{centering}
\begin{tabular}{|c|c|c|c|}
\hline 
Model & IOU & \#Wrong class & \#Wrong label\tabularnewline
\hline 
\hline 
Res50 baseline & 38.8 & 8.2 & 27577\tabularnewline
\hline 
DML-Res50 & \textbf{40.23} & \textbf{4.9} & \textbf{23057}\tabularnewline
\hline 
\end{tabular}
\par\end{centering}
\caption{Results on NYUDv2 dataset. Dense multi-label network has 1.4\% higher IOU and 40\% and 16\% lower wrong classes and labels respectively.}

\label{tab:nyud_exp}
\end{table}

\begin{table}[h]
\begin{centering}
\begin{tabular}{|c|c|}
\hline 
Model & IOU\tabularnewline
\hline 
\hline 
FCN-32s \cite{Long2015} & 29.2\tabularnewline
\hline 
FCN-HHA \cite{Long2015} & 34.0\tabularnewline
\hline 
Context \cite{Lin2016} & 40.0\tabularnewline
\hline 
DML-Res50 (ours) & 40.23\tabularnewline
\hline 
\end{tabular}
\par\end{centering}
\caption{Comparison with other models on NYUDv2 dataset. Our method achieves the best result.}

\label{tab:nyud_comp}
\end{table}

\subsection{Results on SUN-RGBD}
SUN-RGBD \cite{Song2015} is an extension of NYUDv2 \cite{Silberman2012}, which contains 5285 training images and 5050 validation images, and provides pixel labelling masks for 37 classes.

\begin{table}[h]
\begin{centering}
\begin{tabular}{|c|c|c|c|}
\hline 
Model & IOU & \#Wrong class & \#Wrong label\tabularnewline
\hline 
\hline 
Res50 baseline & 39.28 & 5.3 & 24602\tabularnewline
\hline 
DML-Res50 & \textbf{42.34} & \textbf{3.36} & \textbf{20104}\tabularnewline
\hline 
\end{tabular}
\par\end{centering}
\caption{Results on SUN-RGBD dataset. Dense multi-label helps increase the performance by more than 3\% of IOU and decrease the wrong classes and labels by 36\% and 18\%.}

\label{tab:sun_exp}
\end{table}

\begin{figure}[h]
	\begin{center}
		\includegraphics[width=0.9\linewidth]{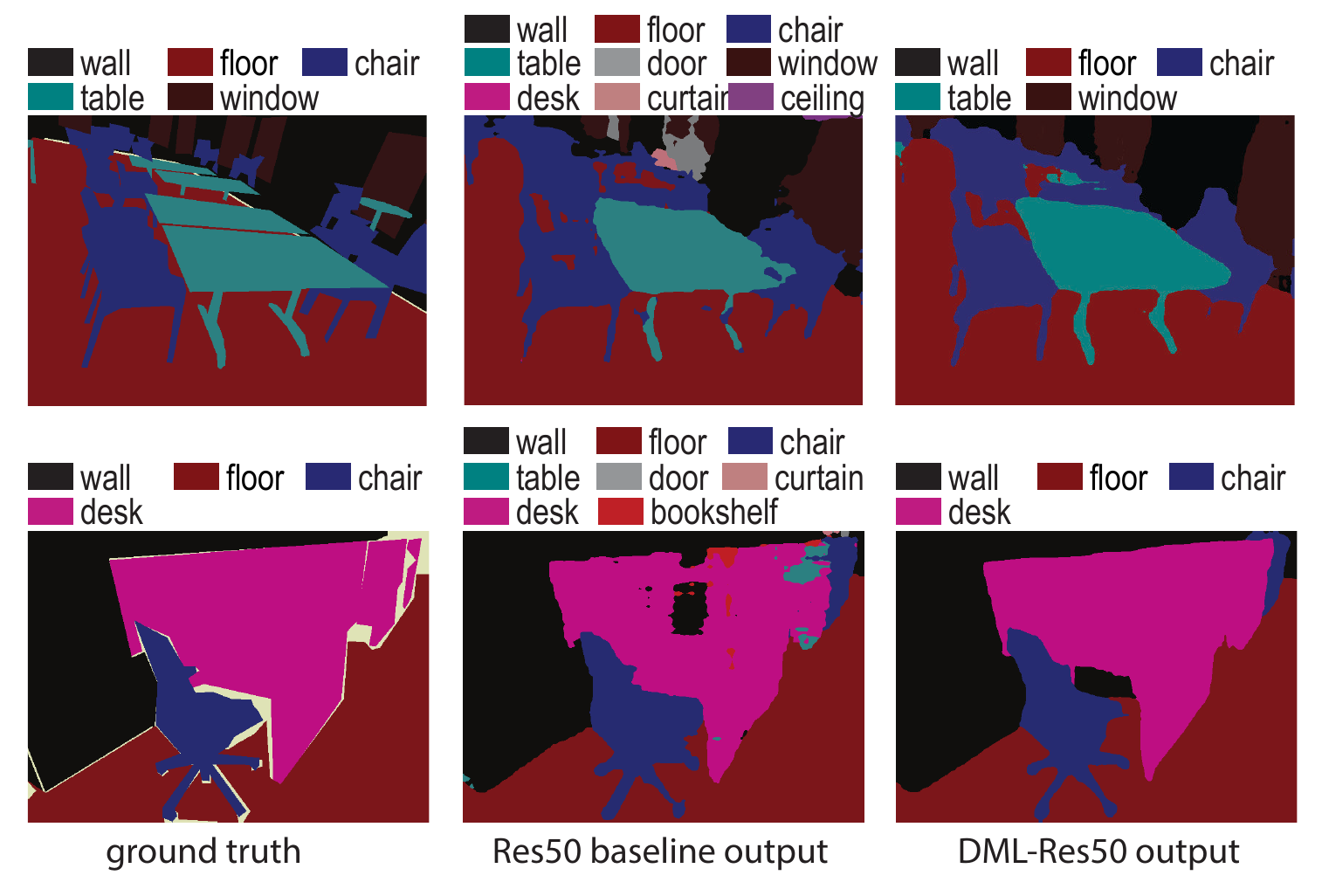}
	\end{center}
	\caption{Example outputs of baseline Res50 and DML-Res50 on SUN-RGBD dataset.}
	\label{fig:sun_exp}
\end{figure}

\begin{figure}[h]
	\begin{center}
		\includegraphics[width=0.9\linewidth]{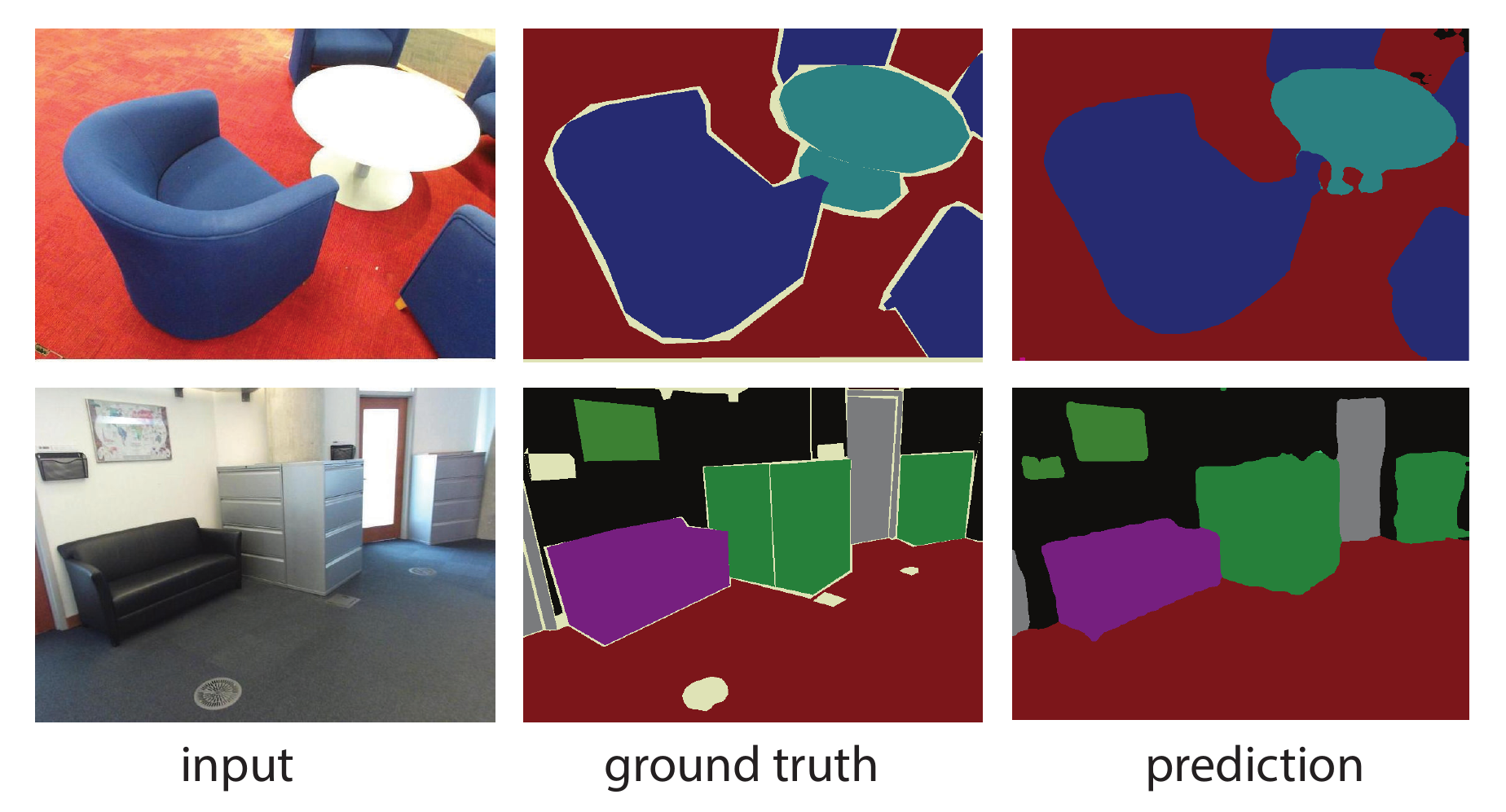}
	\end{center}
	\caption{Good examples on SUN-RGBD dataset.}
	\label{fig:sun_more_exp}
\end{figure}

Figure \ref{fig:sun_exp} shows some output comparison on this dataset, where we can easily observe the effect of dense multi-label.
The results are shown in Table \ref{tab:sun_exp}. The network with dense multi-label helps improve the IOU by more than 3\%. The wrong classes and wrong labels also get decreased by 36\% and 18\% respectively. Compared with other methods, the network with dense multi-label reaches the best result, as shown in Table \ref{tab:sun_comp}. More examples can be found in Figure \ref{fig:sun_more_exp}.
\begin{table}[h]
\begin{centering}
\begin{tabular}{|c|c|}
\hline 
Model & IOU\tabularnewline
\hline 
\hline 
Kendall \etal \cite{Kendall2015} & 30.7\tabularnewline
\hline 
Context \cite{Lin2016} & 42.3\tabularnewline
\hline 
DML-Res50 (ours) & 42.34\tabularnewline
\hline 
\end{tabular}
\par\end{centering}
\caption{Comparison with other models on SUN-RGBD dataset. We achieve the best result with dense multi-label network.}

\label{tab:sun_comp}
\end{table}

\subsection{Ablation Study on PASCAL-Context}

Table \ref{tab:ablation_context} shows an ablation study on the PASCAL-Context. The Res50 baseline yields mean IOU of 41.37\%. Treating this as a baseline, we introduce dense multi-level module. Firstly, in the one level setting, we use the largest window size, which is basically global multi-label classification. Accordding to the results, the first level gives the biggest boost. With 2 levels involved, the global and mid-level window, the performance is improved further. The final level, the smallest window, brings 0.6\% more improvement. The dense multi-label module helps improve the performance by 2.2\% in total. After using CRF as post-processing, we can achieve IOU of 44.39 without using extra MS COCO dataset. 

\begin{table}[h]
\begin{centering}
\begin{tabular}{|c|c|}
\hline 
Model & IOU\tabularnewline
\hline 
\hline 
Res50 baseline & 41.37\tabularnewline
\hline 
DML-Res50 1level  & 42.52\tabularnewline
\hline 
DML-Res50 2level & 42.95\tabularnewline
\hline 
DML-Res50 3level & 43.59\tabularnewline
\hline 
DML-Res50 3level + CRF & \textbf{44.39}\tabularnewline
\hline 
\end{tabular}
\par\end{centering}
\caption{Ablation study on PASCAL-Context.}

\label{tab:ablation_context}
\end{table}

\subsection{Failure Analysis}
\begin{figure}[t!]
\begin{center}
   \includegraphics[width=1.0\linewidth]{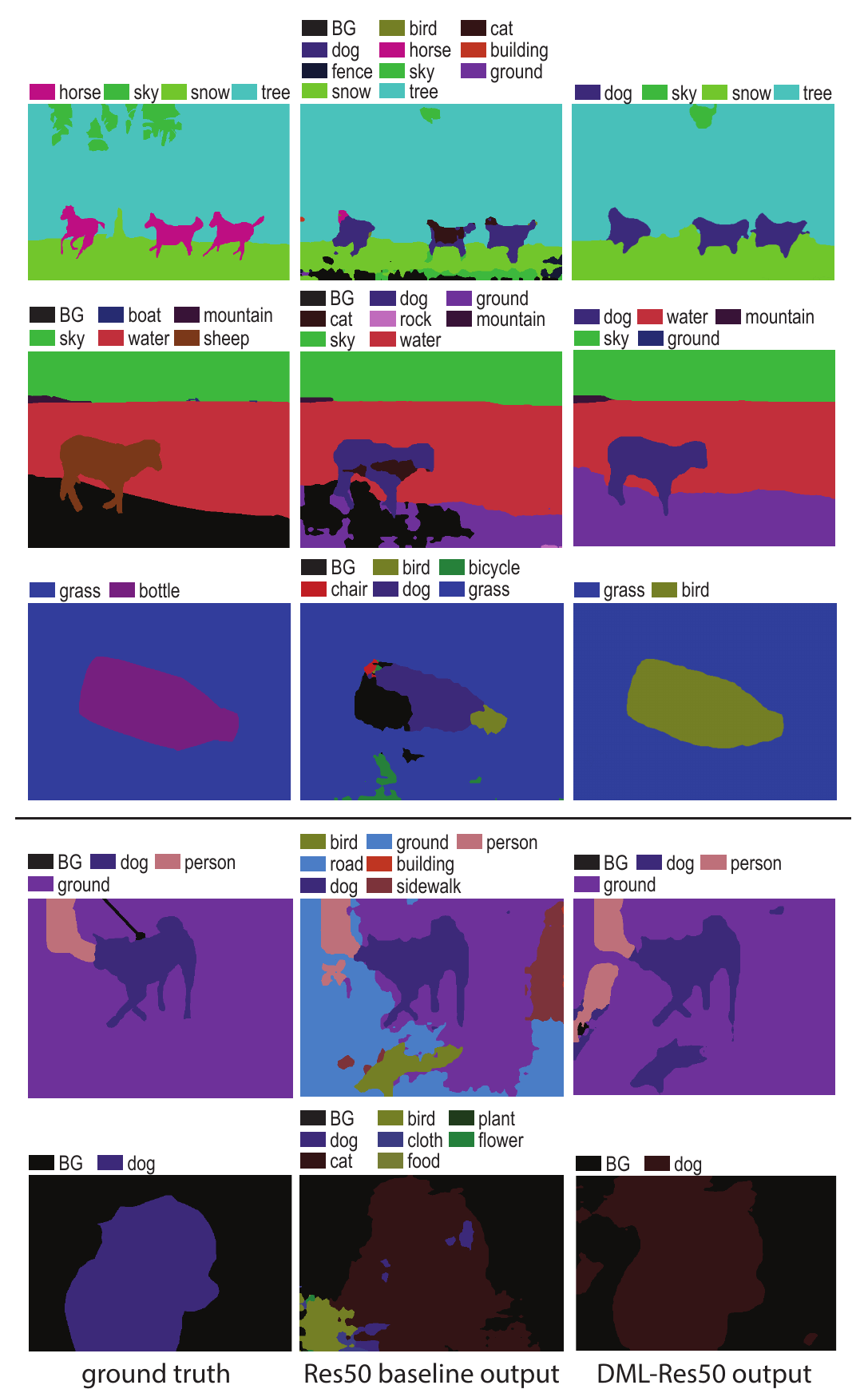}
\end{center}
   \caption{Examples of failed case.}
\label{fig:failed_case}
\end{figure}
We also observed some failure cases from the outputs, with two main types of failure shown in Figure \ref{fig:failed_case}. The left half of Figure \ref{fig:failed_case} depicts a failure mode in which the objects are totally misclassified into another class; here the assigned lables are consistent due to the dense multi-label module but the object/region class is wrong. Another failure type is shown in the right half of the figure, where the labels are consistent but the model failed to detect some objects or detected some non-existing objects.  In the former case, the error here appears primarily to be one exacerbated by the dense multi-label prediction.  This  could be mitigated by improving the quality of dense multi-label prediction and/or adjusting the balance between the dense multi-label module and the segmentation part. We emphasize however, that the dense multi-label technically can be integrated into any segmentation system to help retain the consistency, and our results show the efficacy of doing so.

\section{Conclusion}
\label{conclusion}
In this study, we propose a dense multi-label module to address the problem of scene consistency. With comprehensive experiments, we have shown that dense multi-label can enforce the scene consistency in a simple and effective way. More importantly, the dense multi-label is a module and can be easily integrated into other semantic segmentation systems. 

In terms of future work, we consider investigating better ways to combine the dense multi-label module and segmentation system. In other words, we might conduct research on better methods to fuse the preditions from dense multi-label and segmentation.

{\small
\bibliographystyle{unsrt}
\bibliography{CSRef}

\begin{thebibliography}{10}

\bibitem{Girshick2014}
Ross Girshick, Jeff Donahue, Trevor Darrell, and Jitendra Malik.
\newblock {Rich Feature Hierarchies for Accurate Object Detection and Semantic
  Segmentation}.
\newblock In {\em Proc. IEEE Conf. Comp. Vis. Patt. Recogn.}, pages 580--587,
  2014.

\bibitem{Carreira2012}
Jo{\~{a}}o Carreira, Rui Caseiro, Jorge Batista, and Cristian Sminchisescu.
\newblock {Semantic Segmentation with Second-Order Pooling}.
\newblock In {\em Proc. Eur. Conf. Comp. Vis.}, volume 7578 LNCS, pages
  430--443, 2012.

\bibitem{Hariharan2014}
Bharath Hariharan, Pablo Arbel{\'{a}}ez, Ross Girshick, and Jitendra Malik.
\newblock {Simultaneous Detection and Segmentation}.
\newblock {\em Proc. Eur. Conf. Comp. Vis.}, pages 297--312, 2014.

\bibitem{Yadollahpour2013}
Payman Yadollahpour, Dhruv Batra, and Gregory Shakhnarovich.
\newblock {Discriminative re-ranking of diverse segmentations}.
\newblock {\em Proc. IEEE Conf. Comp. Vis. Patt. Recogn.}, pages 1923--1930,
  2013.

\bibitem{Farabet2013}
Cl{\'{e}}ment Farabet, Camille Couprie, Laurent Najman, and Yann Lecun.
\newblock {Learning Hierarchical Features for Scene Labeling}.
\newblock {\em {IEEE} Trans. Pattern Anal. Mach. Intell.}, 35(8):1915--1929,
  2013.

\bibitem{Cogswell2014}
Michael Cogswell, Xiao Lin, Senthil Purushwalkam, and Dhruv Batra.
\newblock {Combining the Best of Graphical Models and ConvNets for Semantic
  Segmentation}.
\newblock {\em arXiv}, page~13, 2014.

\bibitem{Dai2015}
Jifeng Dai, Kaiming He, and Jian Sun.
\newblock {[M] BoxSup: Exploiting Bounding Boxes to Supervise Convolutional
  Networks for Semantic Segmentation}, 2015.

\bibitem{Hong}
Seunghoon Hong, Hyeonwoo Noh, and Bohyung Han.
\newblock {Decoupled Deep Neural Network for Semi-supervised Semantic
  Segmentation}.
\newblock In {\em Proc. Advances in Neural Inf. Process. Syst.}, pages
  1495--1503, 2015.

\bibitem{Shen2016}
Falong Shen and Gang Zeng.
\newblock {Fast Semantic Image Segmentation with High Order Context and Guided
  Filtering}.
\newblock 2016.

\bibitem{Chen2014a}
Liang-Chieh Chen, George Papandreou, Iasonas Kokkinos, Kevin Murphy, and
  Alan~L. Yuille.
\newblock Semantic image segmentation with deep convolutional nets and fully
  connected {CRFs}.
\newblock In {\em Proc. Int. Conf. Learn. Representations}, 2015.

\bibitem{Lin2015}
Guosheng Lin, Chunhua Shen, Anton {van dan Hengel}, and Ian Reid.
\newblock Efficient piecewise training of deep structured models for semantic
  segmentation.
\newblock 2016.

\bibitem{Chen2016}
Liang-Chieh Chen, George Papandreou, Iasonas Kokkinos, Kevin Murphy, and
  Alan~L. Yuille.
\newblock Deeplab: Semantic image segmentation with deep convolutional nets,
  atrous convolution, and fully connected {CRFs}.
\newblock {\em arXiv: Comp. Res. Repository}, 2016.

\bibitem{Wei2014}
Yunchao Wei, Wei Xia, Junshi Huang, Bingbing Ni, Jian Dong, Yao Zhao, and
  Senior Member.
\newblock {CNN : Single-label to Multi-label}.
\newblock {\em arXiv: Comp. Res. Repository}, abs/1406.5, 2014.

\bibitem{Jiang2016}
Jiang Wang, Yi~Yang, Junhua Mao, Zhiheng Huang, Chang Huang, and Wei Xu.
\newblock {CNN-RNN}: A unified framework for multi-label image classification.
\newblock In {\em Proc. IEEE Conf. Comp. Vis. Patt. Recogn.}, 2016.

\bibitem{Moore2016}
Carl~A Moore, Michael~A Peshkin, and J~Edward Colgate.
\newblock {HCP: A Flexible CNN Framework for Multi-Label Image Classification}.
\newblock {\em {IEEE} Trans. Pattern Anal. Mach. Intell.}, 38(2):1901--1907,
  2016.

\bibitem{Guo2011}
Yuhong Guo and Suicheng Gu.
\newblock {Multi-label classification using conditional dependency networks}.
\newblock In {\em Proc. Int. Joint Conf. Artificial Intell.}, pages 1300--1305,
  2011.

\bibitem{Gong2013}
Yunchao Gong, Yangqing Jia, Thomas Leung, Alexander Toshev, and Sergey Ioffe.
\newblock {Deep Convolutional Ranking for Multilabel Image Annotation}.
\newblock {\em arXiv: Comp. Res. Repository}, pages 1--9, 2013.

\bibitem{Long2015}
Jonathan Long, Evan Shelhamer, and Trevor Darrell.
\newblock {Fully Convolutional Networks for Semantic Segmentation}.
\newblock In {\em Proc. IEEE Conf. Comp. Vis. Patt. Recogn.}, pages 3431--3440,
  2015.

\bibitem{Liu2015b}
Wei Liu, Andrew Rabinovich, and Alexander~C. Berg.
\newblock {ParseNet: Looking Wider to See Better}.
\newblock {\em arXiv preprint: arXiv:1506.04579}, pages 1--11, 2015.

\bibitem{Xue2011}
Xiangyang Xue, Wei Zhang, Jie Zhang, Bin Wu, Jianping Fan, and Yao Lu.
\newblock {Correlative multi-label multi-instance image annotation}.
\newblock {\em Proc. IEEE Int. Conf. Comp. Vis.}, pages 651--658, 2011.

\bibitem{Arge2015}
Karen Simonyan and Andrew Zisserman.
\newblock Very deep convolutional networks for large-scale image recognition.
\newblock In {\em Proc. Int. Conf. Learn. Representations}, pages 1--14, 2015.

\bibitem{Szegedy}
Christian Szegedy, Sergey Ioffe, and Vincent Vanhoucke.
\newblock {Inception-v4, Inception-ResNet and the Impact of Residual
  Connections on Learning}.

\bibitem{Technologii2013}
Kaiming He, Xiangyu Zhang, SHaoqing Ren, and Jian Sun.
\newblock {Deep Residual Learning for Image Recognition}.
\newblock 7(3):171--180, 2013.

\bibitem{Krizhevsky2012}
Alex Krizhevsky, Ilya Sutskever, and Geoffrey~E Hinton.
\newblock {Imagenet classification with deep convolutional neural networks}.
\newblock {\em Proc. Advances in Neural Inf. Process. Syst.}, pages 1106--1114,
  2012.

\bibitem{Everingham2010}
Mark Everingham, Luc {Van Gool}, Christopher K~I Williams, John Winn, and
  Andrew Zisserman.
\newblock {The pascal visual object classes (VOC) challenge}.
\newblock {\em International Journal of Computer Vision}, 88(2):303--338, 2010.

\bibitem{Zhou2016}
Bolei Zhou, Hang Zhao, Xavier Puig, Sanja Fidler, Adela Barriuso, and Antonio
  Torralba.
\newblock {Semantic Understanding of Scenes through the ADE20K Dataset}.
\newblock {\em arXiv}, 2016.

\bibitem{Mottaghi2010}
Roozbeh Mottaghi, Xianjie Chen, Xiaobai Liu, Nam-gyu Cho, Seong-whan Lee,
  Raquel Urtasun, and Alan Yuille.
\newblock {The Role of Context for Object Detection and Semantic Segmentation
  in the Wild}.
\newblock 2010.

\bibitem{Arnab2015}
Anurag Arnab, Sadeep Jayasumana, Shuai Zheng, and Philip Torr.
\newblock {Higher Order Conditional Random Fields in Deep Neural Networks}.
\newblock {\em Arxiv}, page~10, 2015.

\bibitem{Lin2016}
Guosheng Lin, Chunhua Shen, Anton Van~Den Hengel, and Ian Reid.
\newblock Exploring context with deep structured models for semantic
  segmentation.
\newblock {\em Arxiv 2016}, pages 1--14, 2016.

\bibitem{Wu2016b}
Zifeng Wu, Chunhua Shen, and Anton Van~Den Hengel.
\newblock {Bridging Category-level and Instance-level Semantic Image
  Segmentation}.
\newblock 2016.

\bibitem{Lin2014}
Tsung-Yi Lin, Michael Maire, Serge Belongie, Lubomir~D Bourdev, Ross~B
  Girshick, James Hays, Pietro Perona, Deva Ramanan, Piotr Doll{\'{a}}r, and
  C~Lawrence Zitnick.
\newblock Microsoft coco: Common objects in context.
\newblock {\em arXiv:1405.0312}, pages 740--755, 2014.

\bibitem{Silberman2012}
Nathan Silberman, Derek Hoiem, Pushmeet Kohli, and Rob Fergus.
\newblock {Indoor segmentation and support inference from RGBD images}.
\newblock {\em Lecture Notes in Computer Science (including subseries Lecture
  Notes in Artificial Intelligence and Lecture Notes in Bioinformatics)}, 7576
  LNCS(PART 5):746--760, 2012.

\bibitem{Song2015}
Shuran Song, Samuel~P. Lichtenberg, and Jianxiong Xiao.
\newblock {SUN RGB-D: A RGB-D scene understanding benchmark suite}.
\newblock {\em Proc. IEEE Conf. Comp. Vis. Patt. Recogn.}, pages 567--576,
  2015.

\bibitem{Kendall2015}
Alex Kendall, Vijay Badrinarayanan, and Roberto Cipolla.
\newblock {Bayesian SegNet: model uncertainty in deep convolutional
  encoder-decoder architectures for scene understanding}.
\newblock {\em arXiv:1511.02680v1 [cs.CV]}, 2015.

\end{thebibliography}
}

\end{document}